\documentclass[conference]{IEEEtran}
\usepackage{graphicx}
\usepackage{multicol}   

\usepackage{multirow}

\usepackage{cite}

\ifCLASSINFOpdf
\else
\fi
\usepackage[cmex10]{amsmath}
\usepackage{float}
\usepackage{array}
\usepackage{url}

\begin{document}
%
\title{The challenges of SVM optimization using Adaboost on a phoneme recognition problem}


%
\author{\IEEEauthorblockN{Rimah Amami\IEEEauthorrefmark{1},
Dorra Ben Ayed\IEEEauthorrefmark{2} and
Noureddine Ellouze\IEEEauthorrefmark{3}}
\IEEEauthorblockA{Department of Electrical Engineering\\
National School of Engineering of Tunis,
University of Tunis - El Manar, Tunisia}
\IEEEauthorblockA{\IEEEauthorrefmark{1}Email: Rimah.amami@yahoo.fr}}


\maketitle

\begin{abstract}
The use of digital technology is growing at a very fast pace which led to the emergence of systems based on the cognitive infocommunications.  The expansion of this sector impose the use of  combining methods in order to ensure the robustness in cognitive systems.\\ 
Boosting is a technique for combining many weak classifiers to form one high-performance prediction rule  to improve the performance of any given learning algorithm. In theory, Boosting can be used to reduce the error of the learning algorithm which  generates learners that needs to be better than random guessing. Adaboost  is the most used boosting algorithm.
 Support Vector Machines (SVMs) and Adaptive Boosting (Adaboost) are two successful classification methods which are essentially similar as they both try to maximize the minimal margin on a training set.
In this paper, we will investigate the impact of Adaboost on the supervised algorithm SVM for a Multi-Class phoneme recognition problem. This task may be complex since SVM is not an easy classifier to train and so, AdaboostSVM may not be viable.
Furthermore,  we compare the recognition rates to other commonly used Adaboost methods, such as Decision Tree C4.5.
From the different phoneme datasets, we shall show that the single SVM-RBF outperforms Decision Tree C4.5,  the  AdaboostC4.5 and even the Adaboost SVM-Based component classifier.

To experiment AdaboostSVM, we use the phoneme datasets from TIMIT corpus and MFCC feature representations.
\end{abstract}


\section{Introduction}

\label{intro}

Speech recognition proved to be successful in enormous applications in the last few decades. Tthe Cognitive Infocommunications (CogInfoCom) systems  ties in closely with the patterns and speech recognition area since the user and the CogInfoCom systems can interact through a human computer communication \cite{Baranyi} \cite{Sallai}.\\ 
There has been a considerable improvement in speech applications since it is closely related to the information and communications technology and cognitive science problems.

Despite the considerable progress in speech researches, the robustness of speech recognition systems still slight due to many factors such as large data, noisy environments, speaker voice, etc.  Unfortunately, many speech recognition systems have their limits to accomplish the human performance.\\
For example, a phoneme recognition system is based on a sequence of phones but those phones bring the understanding and the confusion with the neighboring phonemes.
Thus, many studies in the speech recognition develop techniques that provide into the speech recognition systems greater levels of knowledge of the language. Indeed, the relationship between speech recognition and cognitive capacity proved to have a direct impact on the human-interaction systems.
But how is the interaction between speech and cognitive science in the development of speech recognition systems?\\

In fact,  the cognitive model based on the speech recognition systems have to better focusing in the descriptions input such as the raw acoustic signal in order to boost the accuracy of the recognition system.
In this context, this paper tries to improve the performance of a speech recognition system based on a raw acoustic input recording from a microphone.
Several speakers (female and male) pronounce few sentences but each speaker has its own behavior, accent, energy, noise, etc.  Those cognitive factors make it difficult to the robustness of the speech recognition system.\\

In the automatic speech recognition (ASR) field, the choice of the learning algorithm for building of any ASR system is a crucial step  since the success of the  recognition task depends essentially on learning stage.
Recently, the Boosting have been  proved to be an efficient method for improving the performance of different classifiers. Therefore, this technique is used to combine a collection of component classifiers (called also weak classifiers), to form a single "strong" classifier. It consists in calling repeatedly the component classifier  on different weights over the training samples and adaptively adjusting these weights after each Boosting iteration \cite{Freundd}.\\

Adaboost is the most popular method of Boosting. The great success of Adaboost can be attributed to its ability to maximize the margin on a training set, which lead to improve the performance of the classifier.
The efficiency of Adaboost is closely related to the component classifier used. Since the main objective of Adaboost is to enhance the learning performance of a given weak classifier,  then  combing Adaboost with a strong classifier may not necessarily be optimum choice and it may be going against the gain of the Boosting principle.
Furthermore, the  choice of the  learning algorithm, generally, affect the recognition rates. But how to distinguish between  a "weak" and a  "strong" learner algorithm ?	\\
We propose this study to emphasize the notion of "the weak and strong  learner"; SVM and C4.5  are two strong algorithms which constitutes our base component classifiers. In particular, we would observe the behaviour of SVM with Adaboost.

The remaining parts of this paper is organized as follows: In section \ref{sec:1}, the main idea of Adaboost is introduced. In section \ref{sec:2}, the learning algorithms used in this study are presented; The architecture of our ASR system is proposed in section \ref{sec:3};  Experimental setup and results are described in sections \ref{sec:4}. The conclusion is made in section \ref{sec:6}.

\section{Adaboost}
\label{sec:1}
The Boosting method was inspired by on-line learning algorithm called Hedge $(\beta)$  which allocate weights to a set of hypothesis used to predict the outcome of any system \cite{Freundd}. Recently, Boosting method has been quite successfully applied on real-world applications and it was mostly used on the speech recognition field. The most and widely used boosting algorithm is AdaBoost.\\
AdaBoost (Adaptive Boosting), an iterative algorithm, was originally introduced by Feund and Schapire on 1995 \cite{Freundd}  \cite{Freundal}.\\
If the applied learning algorithm  had a low performance, Adaboost algorithm  generates a sequentially weighted set of weak classifiers in order to create new classifiers which are more operational on the training data.
Hence, the AdaBoost algorithm  multiple iteratively classifiers  to improve the classification  accuracies of many different data sets compared to the given best individual classifier.\\

The main idea of Adaoost is to run repeatedly a given weak learning algorithm in different probability distributions, $W$,  over the training data.  This distribution is initially set uniform.\\
In the meantime, Adaboost calls the Weak Learner algorithm repeatedly in a series of cycles $T$. Then, it assigns higher weights to the misclassified samples by the current component classifier(at cycle t) , in the hopes that the new weak classifier can reduce the classification error by focusing on it. Meanwhile, lower weights will be assigned  to the correctly classified samples \cite{Donghong} \cite{Haifeng}. Thereafter, the distribution $W$  is updated after each cycle.
In the end, hypothesis produced by the weak learner  from each  cycle are combined  into a single "Strong" hypothesis $f$ \cite{Meiral}.\\
The importance of Adaboost lies in the component classifiers which is systematically have an accuracy slightly higher than 50\%. This means that the component classifiers have to be  better than a random estimation \cite{Li08}.\\

Briefly,  Adaboost is a meta-learning method that tries to build a "strong" learning algorithm based on a group of "weak" classifiers.
It must be pointed out that Boosting has been very successful for solving two-class classification problems.\\
 Meanwhile, to achieve multi-class classification, most algorithms have to convert  the multi-class  problem to a  multiple binary classification problems. Our ASR system will use Adaboost.M1.\\
 
 Since this paper discuss an hybrid of the learning algorithm with Adaboost, we will present an overview of other related studies that used Adaboost in cognitive systems. 
 In the few past years, several studies combining Adaboost and cognitive systems have been developed to take advantage of the Adaboost  algorithm to improve the system's performance.
 Chakraborty present in \cite{Chakraborty} an expert cognitive system which use Adaboost to boosts the performance of an ensemble of classifiers. The empirical comparison of his study shows that hybrid learner based Adaboost outperforms the single weak learner.\\
 Stanciulescu et al. used Adaboost algorithm in order to improve the real-time object detection in complex robotics problem \cite{Stanciulescu}.
 Experimental results of this study  on a car database show that the boosted classifier improve the results of the vehicle-detection application.\\
 On the other hand, Lee and al. used Adaboost for the text detection in natural scene to enhance the detection system \cite{jinlee}.\\
 For a driver's cognitive distraction detection problem, the authors used Adaboost to improve the performance for detection of driver distraction \cite{Miyaji}.
 Based on experimental results, the authors shows the capability of  AdaBoost to enhance the accuracy of a problem  based on the detection of  state of cognitive distraction.\\
 
 Adaboost was combined with SVM for a triaxial accelerometer-based fall detection problem in \cite{WenChang}. The experimental results proves that the proposed method Adaboost-SVM gives optimal results compared to those with the single method. The Adaboost-SVM produces, also,  the the lowest false alarm rate and the detection results as well as the highest accuracy results of their study.\\

\section{Compenent Classifiers}
\label{sec:2}

The principal aim of the learning algorithm is to extract regularities from sets of samples. They are consisting of several algorithms that improve automatically the system through experience \cite{Vapnik}.\\
In this paper, we are interested by the supervised  learning since it allows to generate function that maps inputs to desired outputs.\\
In the last decades, different learning algorithms were applied to identify and verify the speech. There is several algorithms that were mostly used  in the ASR  such as decision tree, LDA, SVM, the baseline system, etc.\\
\subsection{Support Vector Machine }
\label{sec:sub3}

A Support Vector Machine (SVM) is a learning algorithm for pattern recognition and regression problems \cite{Singer} whose approaches the classification problem as an approximate implementation of the Structural Risk Minimization (SRM) induction principle \cite{Cortesal}.\\
 SVM approximates the solution to the minimization problem of SRM through a Quadratic Programming optimization.
It aims to maximize the margin  which is the distance from a separating hyperplane to the closest positive or negative sample between classes \cite{Amamial}.\\

A subset of training samples is chosen as support vectors. They determine the decision boundary hyperplane of the classifier.
Based on this principle, the SVM adopts a systematic approach to find a linear function that belongs to a set of functions with lowest VC dimension (the Vapnik–Chervonenkis dimension measure the capacity of a statistical classification algorithm). \\

Applying a  kernel trick that maps an  input vector into a higher dimensional feature sapce, allows to SVM to approximate a non-linear function \cite{Cortesal} and \cite{Li08} and \cite{Amamial}.
In this paper, we use SVM with the radial basis function kernel (RBF).
 \subsection{Decision Tree C4.5 }
\label{sec:sub4}

 C4.5 is a learning algorithm used to generate a decision tree developed by Ross Quinlan \cite{Quinlan}. The Decision Tree C4.5 is an extension of Quinlan's earlier ID3 algorithm  (Iterative Dichotomiser 3).\\
Just like ID3, C4.5 employs a "divide and conquer" strategy \cite{Kohavial} and uses the concept of information entropy to compute builds decision trees from a set of training (the split criteria)
The  training data is a set $X=x1,x2,…xn$ of  already classified samples. Each sample $X_i=x_1,x_2,\ldots$ is a vector where $x_1,x_2,\ldots$ represent attributes or feature of the sample.\\
The training data is augmented with a vector
$C=c_1,c_2,\ldots where c_1,c_2,\ldots$ represent the class to which each sample belongs \cite{Pomorskial}.\\

At each node of the tree, C4.5 chooses one attribute of the data that most effectively splits its set of samples into subsets enriched in one class or the other. Its criterion is the normalized information gain (difference in entropy) that results from choosing an attribute for splitting the data. The attribute with the highest normalized information gain is chosen to make the decision.\\
With the adaptively boosted C4.5 Decision Tree classifiers \cite{Vapnik}, very high degree of accuracy can be achieved in ASP field.

\section{Phoneme Recognition System Architecture}
\label{sec:3}

\begin{figure*}
  \centering
  \includegraphics[scale=0.6]{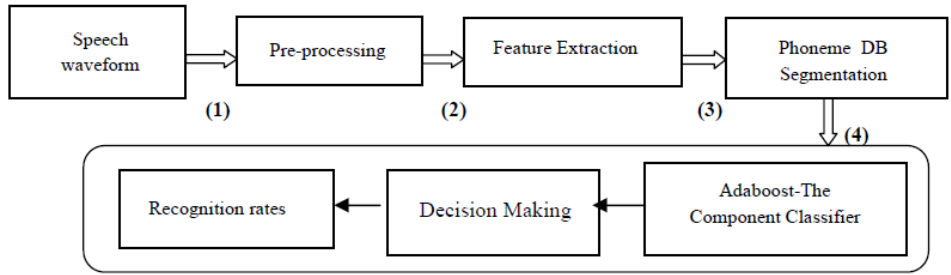}\\
  \caption{Architecture of the ASR system}\label{fig111}
\end{figure*}

 The architecture of our phoneme recognition system is described in this section. The  proposed system aims to compare the performance of Adaboost with SVM-based component classifier and the performance of Adaboost with Decision Tree C4.5-based component classifier on a phoneme recognition task,( see Fig. \ref{fig111}). \\


 \subsection{Adaboost-Component Classifier }
\label{sec:sub5}

The proposed phoneme recognition system passes through several stage with the transformation of the speech samples to spectrogram and then to MFCC spectrum by applying the Spectral  analysis. Then, we proceed to the  the phoneme segmentation to constitute seven sub-phoneme sets. The finale stage consists on applying the proposed algorithms for the phoneme recognition problem, see figure \ref{fig111}.\\

This system aims to use two different algorithms as component classifier in Adaboost. Then, the comparison of error generalization is made in order to see the ability of Adaboost with those two classifiers.

\section{Experimental Setup}
\label{sec:4}

The experiments with our ASR system consists in recognizing which identity of phoneme which been tested (i.e. aa, ae, ih , etc). \\
We split beforehand the phonemes into 7 groups (Vowels,Stops,Nasals,Fricatives,Affricates, Semi-vowels, others(silence))\cite{Amamial}.

As discussed in the previous section, the first setp in a ASR system is the feature extraction. It converts the speech waveform to a set of parametric representation. Hence, we have used the  MEL frequency cepstral coefficients (MFCC) feature extractor.\\
 In 1980,  Davis and Mermelstein developed the MFCC features  for speech classification systems \cite{DavisMermelstein}. It consists on the cepstral coefficients which are produced by the mel-frequency warped Fourier transform function.\\

The system that we present in this paper use the speech samples extracted from the TIMIT corpus \cite{Garofoloal}.
Moreover, for the  nonlinear SVM approach with the “one-against-one” strategy , we choose the RBF (Gaussian) Kernel trick, this choice was made after a previous study done on our data sets with different kernel tricks (Linear, Polynomial, Sigmoid) \cite{Amamial}.\\

The experiments using SVM are done using LibSVM toolbox \cite{ccChang}. The table \ref{tab:tablee1} recapitulate our main choice of experiments conditions:
\begin{table}[H]
  \centering
  \caption{Experimental setup}

\begin{tabular}{|l|l|}
\hline
  & -SVM \\
   \textbf{Methods} & -C4.5\\
  & -AdaboostSVM\\
  & -AdaboostC4.5\\
 \textbf{ $\gamma$} & 1/39 \\
  \textbf{Cost} & 10 \\
  \textbf{Kernel trick} & RBF \\
  \textbf{Windowing} & 3-middle Windows \\ 
  \textbf{Corpus} & TIMIT \\ 
  \textbf{Dialect} & New England \\ 
  \textbf{Frame rate }& 125/s \\ 
  \textbf{Features number} & 39 \\ 
  \textbf{Sampling frequency} & 16ms\\
  \hline
\end{tabular}

  \label{tab:tablee1}%
\end{table}%

\section{Experimental Results }
\label{sec:5}

In this section, SVM and Decision Tree C4.5 are compared with the commonly used Adaboost, which takes SVM and Decision Tree C4.5 as component classifiers. \\

At this stage of experiments, we investigate to find the impact of Adaboost on the two learning classifiers selected.

For the phoneme recognition systems, we compare the performance of our component classifier in Adaboost and single classifiers on 7 data sets which are vowels, semi-vowels, stops, others, nasals, fricatives, affricates.  The final performance of each algorithm on a data set  is the average of the results over the 7 data sets.\\

It must be pointed out that the number of iteration of Adaboost for the recognition system was fixed to 25. In general, the boosted classifier performs well even with only 10 iterations.

\begin{table}[htbp]
  \centering
  \caption{Generalization errors with 3-middle frames of four algorithms per phoneme : SVM-RBF, Adaboost with SVM,  Decision Tree C4.5 and Adaboost with Decision Tree C4.5 }
    \begin{tabular}{l||cccc}
    \hline

    \textbf{Classifiers} & \textbf{SVM} & \textbf{AdaboostSVM} & \textbf{C4.5} & \textbf{AdaboostC4.5}   \\

       \hline
       \hline

    \multirow{2}[1]{*}{\textbf{Vowel}} & \multirow{2}[1]{*}{44.74} & \multirow{2}[1]{*}{45.83} & \multirow{2}[1]{*}{72.95} & \multirow{2}[1]{*}{75.80} \\
         \\
    \multirow{2}[0]{*}{\textbf{Semi-Vowel}} & \multirow{2}[0]{*}{18.55} & \multirow{2}[0]{*}{22.38} & \multirow{2}[0]{*}{38.91} & \multirow{2}[0]{*}{27.62} \\
           \\
    \multirow{2}[0]{*}{\textbf{Stops}} & \multirow{2}[0]{*}{45.72} & \multirow{2}[0]{*}{45.99} & \multirow{2}[0]{*}{64.62} & \multirow{2}[0]{*}{68.45} \\
           \\
    \multirow{2}[0]{*}{\textbf{Others}} & \multirow{2}[0]{*}{14.93} & \multirow{2}[0]{*}{15.97} & \multirow{2}[0]{*}{18.40} & \multirow{2}[0]{*}{16.32} \\
           \\
    \multirow{2}[0]{*}{\textbf{Nasal}} & \multirow{2}[0]{*}{39.46} & \multirow{2}[0]{*}{41.57} & \multirow{2}[0]{*}{61.45} & \multirow{2}[0]{*}{48.80} \\
           \\
    \multirow{2}[0]{*}{\textbf{Fricative}} & \multirow{2}[0]{*}{21.02} & \multirow{2}[0]{*}{26.14} & \multirow{2}[0]{*}{44.51} & \multirow{2}[0]{*}{29.17} \\
            \\
    \multirow{2}[0]{*}{\textbf{Affricate}} & \multirow{2}[0]{*}{21.43} & \multirow{2}[0]{*}{33.33} & \multirow{2}[0]{*}{45.24} & \multirow{2}[0]{*}{33.33} \\
           \\
    \textbf{Average} & 29.40 & 32.95 & 49.44 & 42.78 \\
    \hline
    \end{tabular}%
  \label{tab:tab1}%
\end{table}%

The table \ref{tab:tab1} describes generalization errors of the four algorithms for the given 7 data sets with $3$-middle frames which are Vowel, Semi-Vowel, Fricative, Affricate, Nasal, Silences and Stops .\\

The empirical results present the performance of the phoneme recognition system with the singles classifiers SVM and C4.5 and the combined classifiers. The database is composed by 70\% of learning data and 30\%  of test data which are used for the validation.\\

Since the generalization errors of AdaboostC4.5 are less than those of C4.5 on these 7 data sets. We can conclude that AdaboostC4.5 performs better than C4.5 (i.e. Nasal : 61\% Vs 48\%).\\
In Semi-Vowel, Others, Nasal, Fricative, Affricate, AdaboostC4.5 gets better accuracy  compared with those generates by C4.5 (i.e. the accuracy of Fricative data sets improves about 15\%).\\

In turn, the same conclusion can not be drawn  from table \ref{tab:tab1} for the single SVM-RBF and AdaboostSVM. We observe that the AdaboostSVM performs similarly or worse than the single SVM-RBF. We think that forcing the strong SVM classifiers (SVM with its best parameters) to concentrate on the very hard samples with too much emphasis is the cause of performance degradation in AdaboostSVM algorithm.\\
 This case is also observed in \cite{Wickramaratna} which show that Adaboost with strong classifier component classifier is not effective.\\

For AdaboostSVM, the accuracy have been slightly better on vowel, stops and affricate data sets and the improvement reach about 1,8\%. While, for the rest of data sets used, AdaboostSVM performs slightly worse than single SVM and the declination reach about 4\%.\\

From  table \ref{tab:tab1}, we, also, observed that SVM outperforms both C4.5 and Adaboost C4.5, in general, on these data sets.
Furthermore, for SVM we set a small value of $\gamma$ and also the most suitable for our data ($\gamma$=0.008), which make the SVM classifier stronger. \\
Thus, Adaboost become inefficient because the errors of these component classifiers are highly correlated. Hence, the use of a suitable gamma for SVM component classifier un Adaboost leads to lower the performance of AdaboostSVM for phoneme recognition.\\

We would like to emphasize that the purpose of our experiments is not to argue that SVM-RBF performs better than the decision tree C4.5 and the boosted C4.5, but rather to illustrate that SVM and C4.5 are two strong algorithm, but with SVM optimization problem has been encountered since the performance degradation is the natural expectation of the hybrid Adaboost and the strong SVM-RBF to the multi-class case.\\

In the last decades, some experiments were carried out in order to better understand why boosting sometimes leads to a deterioration in generalization performance. Freund and Schapire put this down to overfitting a large number of  trials T (T for Adaboost  iteration number) which allows to composite classifier to become very complex \cite{Freundal} and \cite{Quinlan1996}. \\

Besides these, boosting try to build a strong classifier from weak classifiers. However, if the performance of the classifier is better than random estimation, then boosting may not result in a strong classifier and this method would be going against the gain of the Boosting principle and not achieve the desired results \cite{Lix}.

\section{Conclusion }
\label{sec:6}
In this paper, we have used Adaboost  as classifier in order to build a phoneme recognition system. We combined Adaboost algorithm with both SVM and C4.5 classifiers in order to show the impact of this algorithm on the performance of recognition system.
For the this  phonemes recognition system we have shown that AdaboostC4.5 have on error rate lower than the single C4.5.  But, single SVM-RBF outperforms AdaboostSVM.\\

Boosting seems to be effective when applied to Decision Tee C4.5 than SVM-RBF although the performance  of the adaboosted C4.5- based component classifier is less than  single SVM. \\
Hence, boosting can not improve the performance of SVMs, and we guess the reason of this phenomenon is that SVMs is a strong classifier. We also found Adaboost with SVM always achieves worse results on all of the phoneme collections.\\
The future work will be to extend our work in order to improves the phoneme recognition accuracy.

\end{document}